\documentclass[UTF8]{nature}
\usepackage{color}
\usepackage{graphicx}
\usepackage{caption}
\usepackage{multicol}
\usepackage{lineno}
\usepackage{ulem}
\title{A new direction to promote the implementation of artificial intelligence in natural clinical settings}

\author{Yunyou Huang$^{1,2,5}$, Zhifei Zhang$^{3,6}$, Nana Wang$^{2,5}$, Nengquan Li$^{4,5}$, Mengjia Du$^{1,2}$, Tianshu Hao$^{1,2}$ \& Jianfeng Zhan$^{1,2,6}$}

\begin{document}
%\linenumbers
\captionsetup[figure]{labelfont={bf},labelformat={default},labelsep=period,name={Fig.}}
\maketitle

\begin{affiliations}
 \item State Key Laboratory of Computer Architecture, Institute of Computing Technology, Chinese Academy of Sciences, Beijing, China.
 \item University of Chinese Academy of Sciences, Beijing, China.
 \item Department of Physiology and Pathophysiology, Capital Medical University, Beijing, China.
 \item Shenzhen University, Shenzhen, Guangdong, China.
 \item These authors contributed equally to this work.
 \item e-mail: zhifeiz@ccmu.edu.cn, e-mail: zhanjianfeng@ict.ac.cn
 \end{affiliations}

\begin{abstract}
Artificial intelligence (AI) researchers claim that they have made great `achievements' in clinical realms. However, clinicians point out the so-called `achievements' have no ability to implement into natural clinical settings. The root cause for this huge gap is that many essential features of natural clinical tasks are overlooked by AI system developers without medical background. In this paper, we propose that the clinical benchmark suite is a novel and promising direction to capture the essential features of the real-world clinical tasks, hence qualifies itself for guiding the development of AI systems, promoting the implementation of AI in real-world clinical practice.
\end{abstract}
\vskip 18pt

AI researchers claim that they have obtained many significant `achievements' in various realms of clinical medicine, i.e., cancer diagnosis~\cite{yu2018artificial, koch2018artificial}. However, in practice, most of the AI products fail to obtain approval from the Food and Drug Administration (FDA). Moreover, the approved AI products, which are quite rare, are only limited to class \uppercase\expandafter{\romannumeral2}  or \uppercase\expandafter{\romannumeral1}\footnote{Medical devices are classified into one of three classes based on the degree of risk they present ~\cite{FDA_2}. The instances for class \uppercase\expandafter{\romannumeral1}, \uppercase\expandafter{\romannumeral2} and \uppercase\expandafter{\romannumeral3} are manual toothbrush, male condoms and heart valve respectively.}, which means that even the approved AI devices are not qualified handling high-risk tasks such as clinical diagnosis\footnote{Since the clinical diagnosis is the base or evidence to the treatment or further action of patient, the incorrect diagnosis may lead to serious consequences.}. Question marks hang over the AI systems for real-world clinical tasks.
Why is there such a huge gap between the AI research and AI implementation in natural clinical setting?
How to promote the AI implementation into natural clinical settings to bridge the gap?

The common interpretation, given by the clinicians, for this huge gap is that many technical issues in clinical settings remain unsolved, leading to the inability of the AI system in natural clinical settings, and that the AI researchers
overestimate the ability of AI system validated in the artificially designed experiments~\cite{vollmer2018machine,miller2018artificial}.
In order to uncover the essential reasons for the gap mentioned above, our team, consisting of AI researchers and clinicians, analyzed the development process of AI systems and give the following explanation.
The features of clinical tasks in natural settings are ignored in the entire lifecycle (design, implementation, and evaluation) of the AI systems, thus the generated AI system itself has no ability to be implemented in natural clinical settings. For example, when the related diseases with similar symptoms are omitted, the AI system cannot directly handle the patients in natural clinical settings\footnote{The patients with the omitted diseases usually appear in the clinical settings.}. Another example is that when the natural ratio between disease-positive and disease-negative subjects is omitted, a risk of spectrum bias and thus the inflated diagnostic or predictive performance will be brought about~\cite{park2018methodologic}.
Thus, it is imperative to capture the essential features of the real-world clinical tasks to guide the entire lifecycle of the AI system and promote its implementation and wide adoption in natural clinical settings.

In the computer community, a domain or application-specific benchmark suite serves as a highly valuable tool to capture real features of the tasks, simulate the tasks in real world, guide the development process, and finally evaluate the system under test.
Inspired by its great success, we strongly believe that the clinical benchmark suite holds the power to improve the design and implementation of AI systems and enables the AI researches to smoothly `touch down' in the real clinical practice.
In the future, the clinical benchmark suite built with the joint efforts of the AI researchers and clinicians will play a significant role in the whole development process of the AI systems, and promote the implementation of AI systems in real clinical settings. The clinical benchmark suite will run throughout the entire lifecycle of the AI system to promote the embedment of AI systems in the current healthcare system. In this way, the AI system will be able to reduce the workload of clinicians, improve the skill of junior clinicians, give precaution and inform the patients to seek the medicine services further in hospitals, and provide preliminary healthcare service in the rural areas.

%give precaution and suggest further medicine service in hospital for patients
\vskip 12pt
\noindent\textbf{State-of-the-art and state-of-the-practice in clinical settings.}

In general, AI technologies went through three periods: expert system, traditional machine learning, and deep learning~\cite{ ghahramani2015probabilistic, berner1994performance, kononenko2001machine}. However, AI systems have hardly been implemented in natural clinical settings, in spite of their plentiful great results achieved in research.

%although they have achieved many great results in researches.

\vskip 12pt
\noindent\textbf{State-of-the-art.} 
 AI technologies have achieved many `achievements' in various realms of clinical medicine such as cancer~\cite{esteva2017dermatologist,capper2018dna,bejnordi2017diagnostic}, cardiovascular diseases~\cite{hannun2019cardiologist,samad2018predicting,tison2018passive} and retinopathy~\cite{gulshan2016development,kermany2018identifying,de2018clinically}.
 In a natural clinical setting, each clinical diagnosis task is flexible, and usually consists of several different sub-tasks according to each individual  patient.
Currently, most reported AI systems outperform the human experts only in certain specific sub-tasks, especially in the medicine imaging area. For example, Esteva et al.~\cite{esteva2017dermatologist} developed a deep convolutional neural network model to classify skin lesions in a task, which was divided into two sub-tasks. One sub-task was to classify the skin lesions into three categories: benign lesions, malignant lesions, and non-neoplastic. As a result, the system defeated the dermatologists (accuracy of $65.78\pm0.22\%$) with an accuracy of $72.1\pm0.9\%$. The other sub-task was to classify the skin lesions into nine categories, the patients of which  have similar medical treatment plans. In this sub-task, the model defeated the dermatologists (accuracy of $54.15\pm0.85\%$) with an accuracy of $55.4\pm1.7\%$.

\vskip 12pt
\noindent\textbf{State-of-the-practice.}
We discuss the state-of-practice from two aspects: the listing qualification and the application performance in a real environment.

\newcommand{\tabincell}[2]{\begin{tabular}{@{}#1@{}}#2\end{tabular}}

\begin{table*}
\footnotesize
\caption{AI medical devices approved  by FDA.\label{Tab1}}
\begin{center}
\scalebox{0.78}[0.78]{
\begin{tabular}{|c|c|c|c|c|clclp{0.5\columnwidth}|}
\hline
\textbf{Approval Date} & \textbf{Device} & \textbf{Function}  & \textbf{\tabincell{c}{Regulatory \\Class}} & \textbf{\tabincell{c}{Submission \\Type}} \\ \hline

December 2018 & ProFound AI & \tabincell{c}{Detection and diagnosis\\ of suspicious lesions} & \uppercase\expandafter{\romannumeral2} & 510(k) \\ \hline
December 2018 & ReSET-O & \tabincell{c}{Adjuvant treatment of \\substance abuse disorder} & \uppercase\expandafter{\romannumeral2} & 510(k) \\ \hline
December 2018 & Embrace & Epilepsy detection & \uppercase\expandafter{\romannumeral2} & 510(k) \\ \hline
November 2018 & Icobrain & Brain structure interpretation & \uppercase\expandafter{\romannumeral2} & 510(k) \\ \hline
October 2018 & Accipiolx & \tabincell{c}{Acute cerebral \\hemorrhage diagnosis} &\uppercase\expandafter{\romannumeral2} & 510(k) \\ \hline
September 2018 & \tabincell{c}{Irregular Rhythm\\ Notification Feature} & Atrial fibrillation detection & \uppercase\expandafter{\romannumeral2} & De Novo \\ \hline
September 2018 & \tabincell{c}{RightEye \\Vision System} & \tabincell{c}{Identify visual \\tracking impairment} & \uppercase\expandafter{\romannumeral2} & 510(k) \\ \hline
August 2018 & BriefCase & \tabincell{c}{Prioritization, \\triage and diagnosis of \\time sensitive patient} & \uppercase\expandafter{\romannumeral2} & 510(k) \\ \hline
August 2018 & \tabincell{c}{PhysiQ Heart \\Rhythm Module} & \tabincell{c}{Atrial fibrillation detection\\ and analysis engine} & \uppercase\expandafter{\romannumeral2} & 510(k) \\ \hline
June 2018 & POGO & \tabincell{c}{Automatic blood \\glucose monitoring system} & \uppercase\expandafter{\romannumeral2} & 510(k) \\ \hline
June 2018 & HealthCCS & \tabincell{c}{Coronary artery calcification\\ detection and quantification} & \uppercase\expandafter{\romannumeral2} & 510(k) \\ \hline
June 2018 & \tabincell{c}{DreaMed \\Advisor Pro }&  \tabincell{c}{Suggestions for parameter \\adjustment of insulin pump }& \uppercase\expandafter{\romannumeral2} & De Novo \\ \hline
May 2018 & NeuralBot & \tabincell{c}{Transcranial \\Doppler probe positioning} & \uppercase\expandafter{\romannumeral2} & 510(k) \\ \hline
May 2018  & MindMotion GO & Motion capture for the elderly & \uppercase\expandafter{\romannumeral2} & 510(k) \\ \hline
May 2018 & EchoMD AutoEF &  \tabincell{c}{Left ventricular \\ejection fraction estimation} & \uppercase\expandafter{\romannumeral2} & 510(k) \\ \hline
May 2018 & OsteoDetect & \tabincell{c}{Identification of \\distal radius fractures} & \uppercase\expandafter{\romannumeral2} & De Novo \\ \hline
April 2018 & IDX-DR & Screening for diabetic retinopathy & \uppercase\expandafter{\romannumeral2} & De Novo \\ \hline
Feburary 2018 & ContaCT & \tabincell{c}{LVO large blood \\vessel blockage warning} & \uppercase\expandafter{\romannumeral2} & De Novo \\ \hline
January 2018 & Embrace & Epilepsy detection & \uppercase\expandafter{\romannumeral2} & 510(k) \\ \hline
January 2018 & \tabincell{c}{Arterys \\Oncology DL} & \tabincell{c}{Volumetric segmentation of \\lung nodules and liver lesions} & \uppercase\expandafter{\romannumeral2} & 510(k) \\ \hline
December 2017 & BioFlux & Arrhythmia detection & \uppercase\expandafter{\romannumeral2} & 510(k) \\ \hline
November 2017 & \tabincell{c}{Kardia \\Band System} & Cardiac arrhythmias detection & \uppercase\expandafter{\romannumeral2} & 510(k) \\ \hline
July 2017 & QuantX  & \tabincell{c}{Assessment and characterization\\ of breast abnormalities} & \uppercase\expandafter{\romannumeral2} & De Novo \\ \hline
May 2017 & AmCAD-US & \tabincell{c}{Visualization and quantitative \\analysis of thyroid nodules }& \uppercase\expandafter{\romannumeral2} & 510(k) \\ \hline
March 2017& EnsoSleep &  Diagnosis of sleep disorders & \uppercase\expandafter{\romannumeral2} & 510(k) \\ \hline
Feburary 2017 & \tabincell{c}{Arterys \\Cardio DL }& \tabincell{c}{Visualization of cardiac structure,\\ quantification of cardiac function} &\uppercase\expandafter{\romannumeral2} & 510(k) \\ \hline
January 2017 & Cantab Mobile & \tabincell{c}{Memory \\problem assessment\\ tool for the elderly} & \uppercase\expandafter{\romannumeral2} & 510(k) \\ \hline
November 2016 & \tabincell{c}{One Drop\\ Blood Glucose \\Monitoring System} & \tabincell{c}{Quantification of\\ blood glucose levels} & \uppercase\expandafter{\romannumeral2} & 510(k) \\ \hline
October 2016  & Lumify & \tabincell{c}{Ultrasound image diagnosis\\ and fluid flow analysis} & \uppercase\expandafter{\romannumeral2} & 510(k) \\ \hline
July 2016 & InPen &  Determine the insulin dose & \uppercase\expandafter{\romannumeral2} & 510(k) \\ \hline
March 2016  & QbCheck & \tabincell{c}{Diagnosis \\and treatment of ADHD} & \uppercase\expandafter{\romannumeral2} & 510(k) \\ \hline
\end{tabular}
}
\end{center}
\end{table*}

In contrast with the plentiful successful AI researches on the clinical tasks, there are few AI products that have obtained FDA approval (Table~\ref{Tab1})~\cite{non2018,topol2019high}, though FDA (the Listing qualification institution) downgraded its regulatory requirements for AI product approval.
 What's more, the existing AI products with FDA approval are incompetent to provide the evidence or basis for the decision of the treatment of the serious diseases for the following reasons.
\begin{itemize}
\item[a)] Most of the incorrect decisions of the treatment may cause the enormous loss to the patients.
\item[b)] All approved AI products, which are only classified into the class \uppercase\expandafter{\romannumeral2}  or \uppercase\expandafter{\romannumeral1}, are not qualified for handling the tremendous risk, since it is stipulated by FDA that medical devices with high risk should be classified into class \uppercase\expandafter{\romannumeral3}  and subject to the strictest supervision.
\end{itemize}
For example, although the first autonomous AI diagnostic system IDx-DR authorized by the FDA is able to give suggestions to a patient after diagnosis~\cite{abramoff2018pivotal}, the suggestions\footnote{The incorrect suggestion may lead to delay of further treatment and loss of vision, which is an enormous risk that the medical device with class \uppercase\expandafter{\romannumeral2} is not qualified to handle. } cannot apply to guide the treatment according to the FDA. What's more, none of the successful AI researches on clinical tasks have obtained the approval from China Food and Drug Administration (CFDA), despite the fact that the Chinese government is eager to use AI to improve the national healthcare.

In the real clinical practice, the AI technologies are far from satisfactory. For example, it was pointed out that many of recommendations for treatment from the Watson for Oncology, the most famous AI product, were erroneous~\cite{topol2019high}. In a randomized controlled trial, Brocklehurst et al.~\cite{brocklehurst2017computerised} reported that the widely-used continuous electronic fetal heart rate monitoring technology did not reduce the number of poor neonatal outcomes in the real clinical practice. Kanagasingam et al.~\cite{kanagasingam2018evaluation} evaluated an AI system serving to identify diabetic retinopathy in clinical practice. As a result, the positive predictive value was 12\% (95\% CI, 8\%-18\%), though the specificity was 92\% (95\% CI, 87\%-96\%).

Based on the above analysis, it has to be admitted that the AI is currently seeing too much hype, and a strict evaluation system is urgently needed to evaluate the clinical value of currently AI system in real clinical practice~\cite{chen2017machine, miller2018artificial, gyawali2018does, fogel2018artificial, goldhahn2018could, Jianxing2019, topol2019high, maddox2019questions}.
%It has to be admitted that the AI hypes too much, and the strict evaluation system is needed
\vskip 12pt
\noindent\textbf{The interpretation for the plight of practical implementation of AI.}

\noindent Currently, the AI systems are developed and validated based on the artificially constructed datasets. However, the natural clinical setting is open, containing many unpredictable factors. Thus, there are many technical issues worth considering before the practical implementation of AI, which is consistent with the views of  the clinicians.

The technical issues, which were proposed by clinicians, are roughly divided into five categories~\cite{anderson2017clinicians,vollmer2018machine}: \textbf{A}) the uncertainty and complexity of clinical tasks in natural clinical settings, \textbf{B}) the metrics of the benefit of the patients, \textbf{C}) the evaluation baselines of the AI systems, \textbf{D}) the reproducibility of the AI system. \textbf{E}) the interpretability of the AI systems. Among them, the issues of reproducibility and interpretability attracted intensive attention from the AI researchers~\cite{ribeiro2016should}, with many significant achievements~\cite{lee2018explainable,kermany2018identifying}. In contrast, the first three types of issues (categories \textbf{A}, \textbf{B}, and \textbf{C}) received relatively less research attention, and have been the cruxes to the implementation of the AI researches into natural clinical settings.

For example, Esteva et al.~\cite{esteva2017dermatologist} claimed that the AI system (a deep convolutional neural network model) defeated the dermatologists in the task of classifying skin lesions, while Mar et al.~\cite{mar2017computer} pointed out that the AI system still needed to solve many issues before it cam be put into practical application. The main items are as follows: 1)  Integrating patient information (\textbf{Category A}), i.e., the history of the lesion, to improve the performance of the AI system. 2) Identifying skin cancers  the patient is unaware of\footnote{The AI model only has the ability to identify the cancers which are contained in train set. Usually, the AI researcher will carefully construct the train set, and the train set only contains a few numbers of diseases.} (\textbf{Category A}). 3) Validating the AI system in a natural clinical setting (\textbf{Category C}). 4) Proving the benefit of the AI system (\textbf{Category B}), such as the reduction in melanoma mortality.
In addition, Adamson et al.~\cite{adamson2018machine} pointed out another issue to be addressed for the AI system: it should hold the inclusivity to handle different types of skin (\textbf{Category A}), i.e., skins with different colors.

 The issues proposed by the clinicians urge us to uncover the root cause for the current plight of the practical implementation of AI to promote its implementation in natural clinical settings.

\vskip 12pt
\noindent\textbf{A new reason for the plight of the practical implementation of AI.}

\noindent In order to uncover the root cause for the plight of AI in natural clinical settings, we successively analyze and summarize the development process of a specific AI system~\cite{esteva2017dermatologist}, the development process of a general AI system,  and the clinical task in a natural clinical setting.

\vskip 12pt
\noindent\textbf{The development process of a specific AI system.} In the design, implementation, and evaluation of the AI system on skin cancer classification by Esteva et al.~\cite{esteva2017dermatologist},  the features of natural clinical setting were omitted as follows in order to simplify the clinical task.
\begin{itemize}
\item[a)] Design: 1) The types of disease of patients are limited. For example, the epidermal lesion must belong to one of the following diseases, malignant basal and squamous cell carcinomas, intraepithelial carcinomas, pre-malignant actinic keratosis and benign seborrheic keratosis. 2) The comorbidity was ignored by assigning only one label for an individual patient. It means that the patient has to obtain an exact primary diagnosis before getting access to the AI system.
%, and according to the primary diagnosis the proper patient will sent to the AI system for further diagnosis.
\item[b)] Implementation: Only image was selected as the patient information. Though the accuracy of the classification of skin lesions was only $55.4\pm1.7\%$,  Esteva et al.~\cite{esteva2017dermatologist} did not integrate other relevant information, such as history of the lesion, to improve the accuracy of the AI system.
\item[c)] Evaluation: Issues regarding the evaluation of the AI system are as follows: 1) The AI system was validated on a carefully constructed dataset, which ignored the features of a natural clinical setting. For example, a number ratio of $169:207$ for malignant vs. benign melanocytic lesions was used in a test dataset, while malignant melanocytic lesions are exceedingly rare compared with benign melanocytic lesions in reality. 2) An unfair baseline was actually constructed. The skin lesions diagnosis is an open task in a natural clinical setting, in which the dermatologists will do their best (even including biopsy) to obtain a diagnosis as accurate as possible for each individual patient. Thus, it is unfair for a clinician to make a diagnosis only based on images. 3) The benefit of patients were ignored. Esteva et al.~\cite{esteva2017dermatologist} did not prove the benefit, such as the reduction in melanoma mortality, of the patients.
\end{itemize}

\begin{figure*}
\centering
\includegraphics[height=6cm, width=10cm]{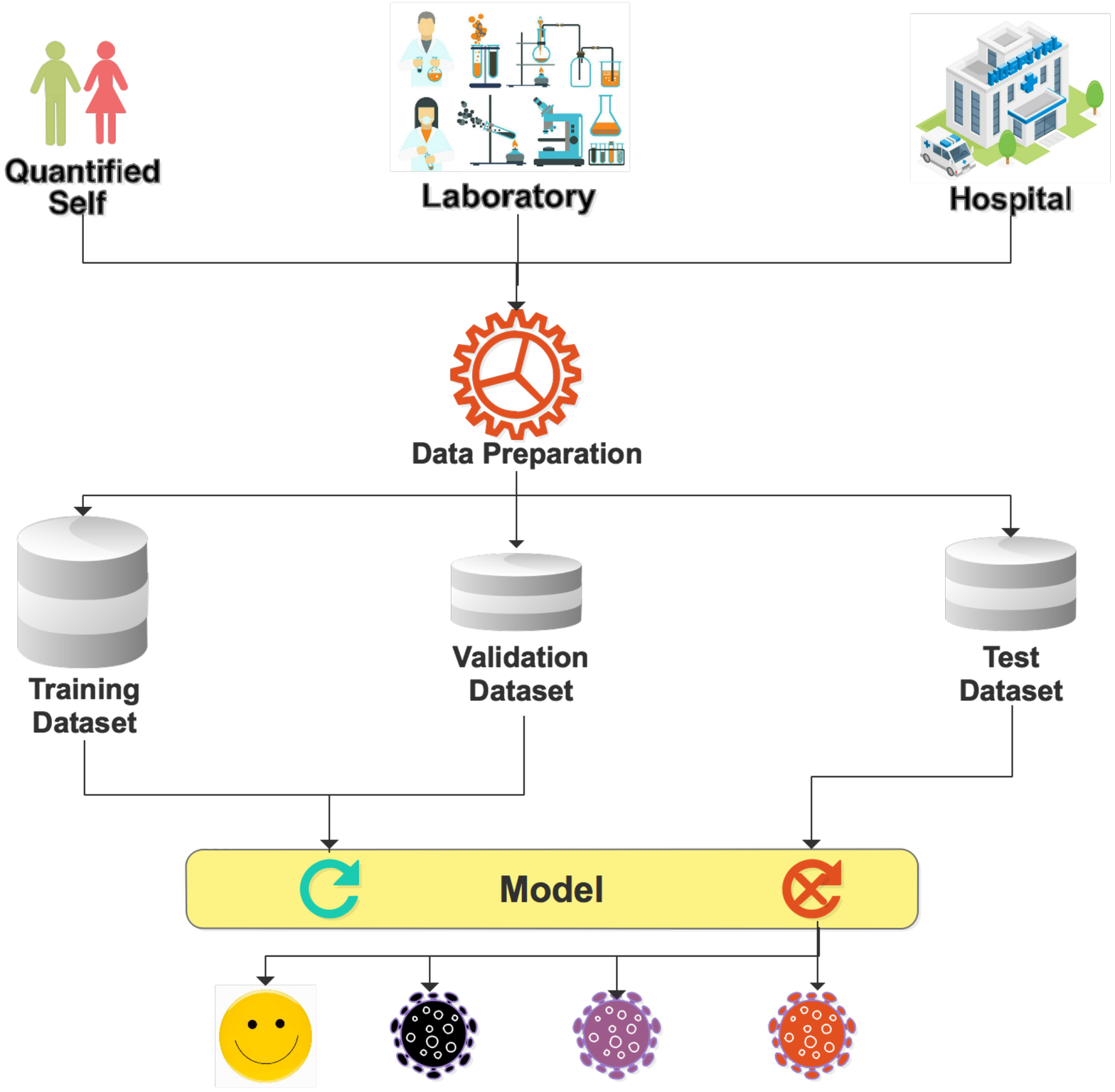}
\quad
\caption{\textbf{The development workflow of an AI clinical system.} For the AI system, first, the researchers collect the relevant information of a clinical diagnosis task from a variety of sources. Second, the data are preprocessed for the model, which includes data selection, data cleaning, data enhancement et al. Finally, the data are divided into a training data set, a validation data set, and a test data set. The training data set and validation data set are used to recurrently train and select the best model, and the test data set is used to evaluate the AI system or AI model. \label{Fig1}}
\end{figure*}

Based on the above analysis. it is evident that many important features of the real-world clinical tasks are omitted during the development of this specific AI system, which makes it difficult to be implemented in the complex and uncertain clinical settings, arousing the aforementioned issues concerned by many clinicians.
\vskip 12pt
\noindent\textbf{The development process of a general AI system.} In order to demonstrate the universal existence of the concerned issues of the state-of-the-art AI systems in a natural clinical setting, we herein discuss a clinical diagnosis task in which a general AI system was used without considering a specific disease.  Currently, the AI researchers usually simplify a specific clinical diagnosis task in the stage of data collection and data preparation (Fig~\ref{Fig1})~\cite{chen2017machine} for the AI system to achieve excellent performance. The issues are as follows.
\begin{itemize}
\item[a)] The task is simplified by the AI developers. The following facts can be observed in the state-of-the-art researches.  1)Excluding related disease that co-occurs with the target disease. 2)Ignoring the comorbidity. 3)Ignoring the so-called `bad data'.

\item[b)] The patient data are carefully selected by the AI researchers, which ensures that each patient has the same type of data, meaning that the patients have the same diagnosis procedure before the therapy. Besides, the selected data usually are incomplete and hardly represent the full health status of a patient.

\item[c)] The evaluation is usually performed in a well-designed and restricted environment.
\end{itemize}
Apparently, many important features are omitted by the AI
researchers, i.e.,  the natural ratio between disease-positive and disease-negative~\cite{park2018methodologic}, during the development of the AI system.

\begin{figure*}
\centering
\begin{minipage}[a]{0.35\linewidth}
\leftline{a}
\includegraphics[height=4.47cm, width=5.44cm]{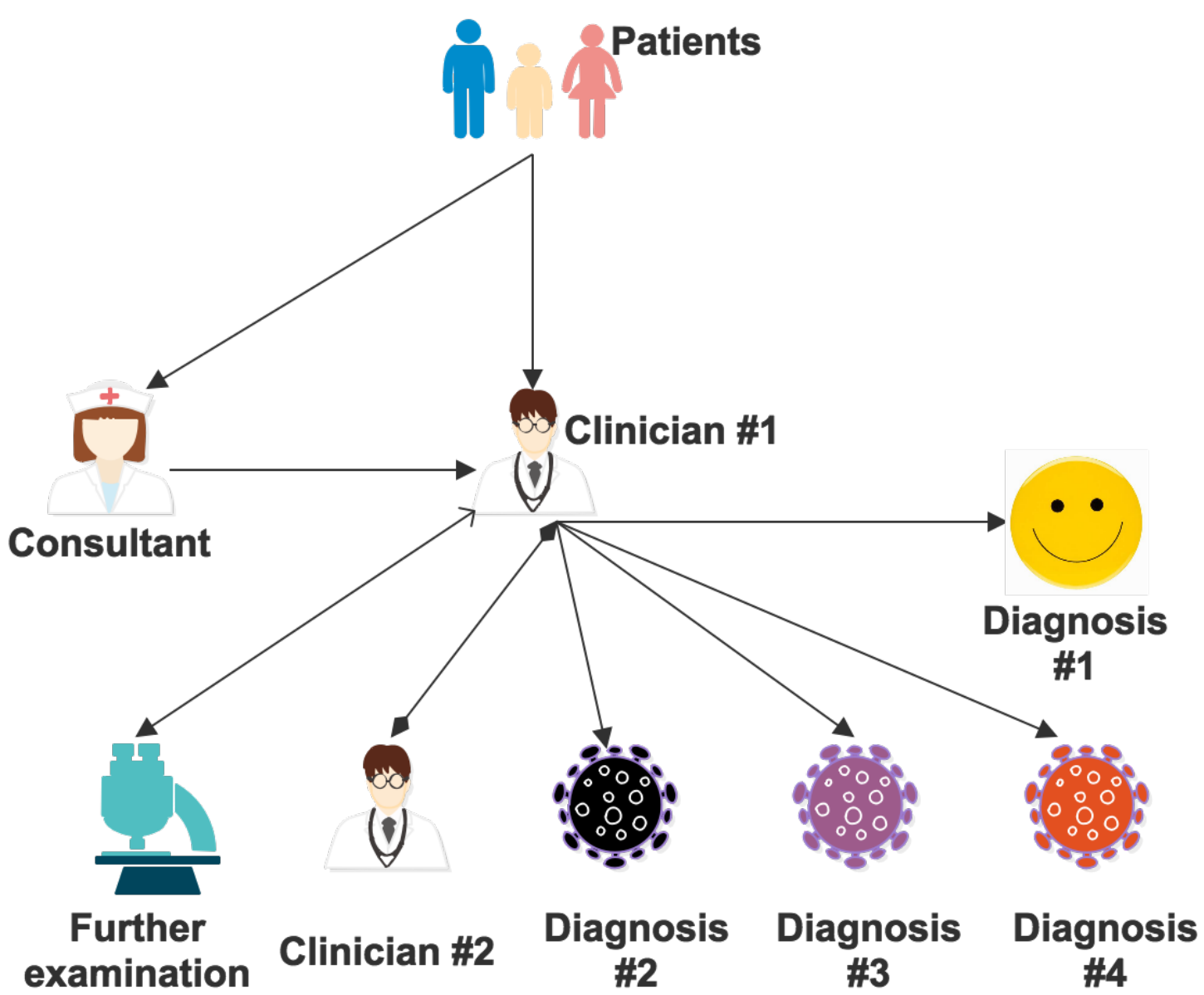}
\end{minipage}
\hspace{0.5cm}
\begin{minipage}[a]{0.55\linewidth}
\leftline{b}
\includegraphics[height=9.02cm, width=9.42cm]{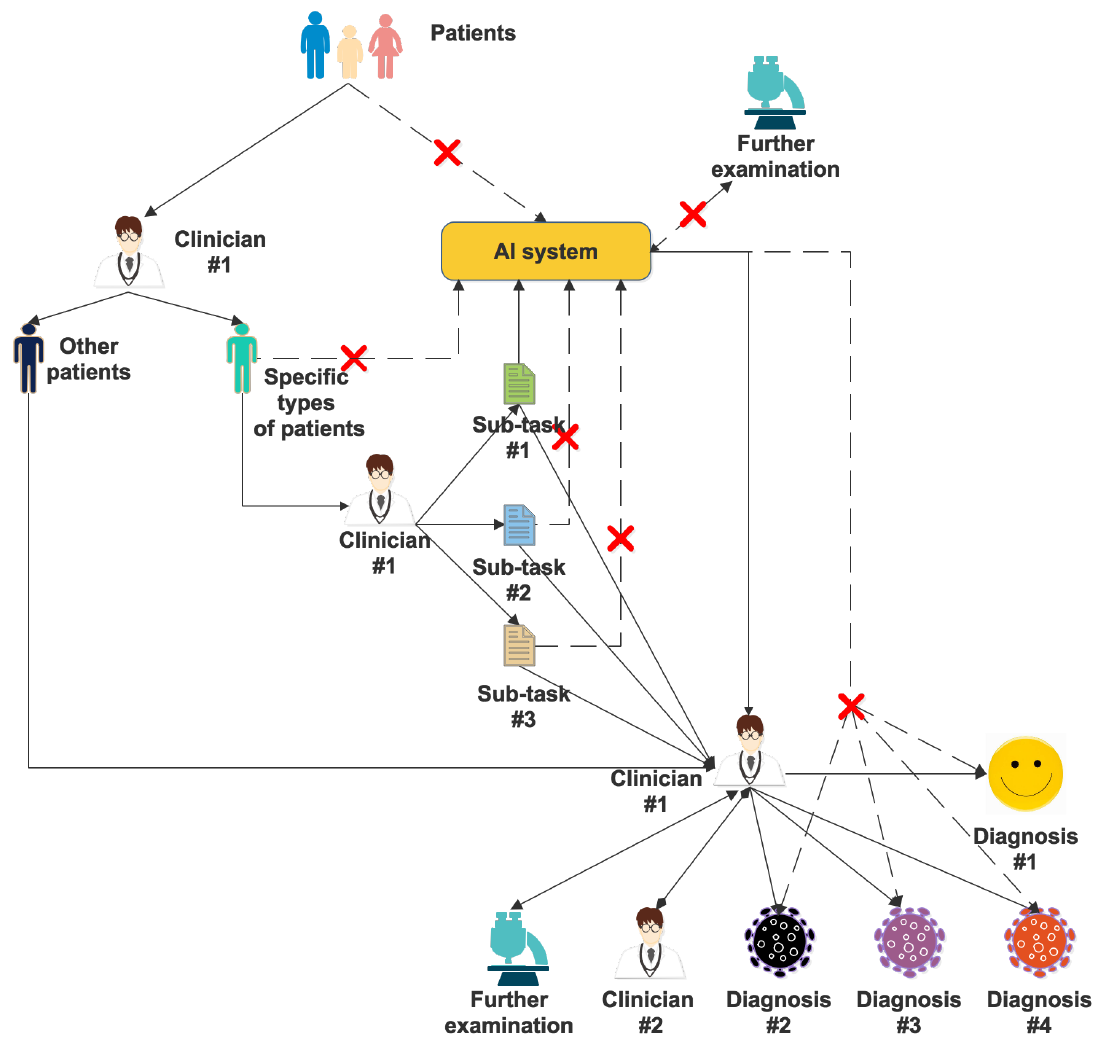}
\end{minipage}\\
\quad
\caption{\textbf{The classical workflow of clinical diagnosis vs. the workflow of clinical diagnosis with current AI systems, which is rigid and disconnected with clinician-centered healthcare systems. a,} In a natural clinical setting, first, a patient visits the clinician according to their experience or the suggestion of the guide nurses. Second, the clinician acquires the necessary information from the patient. Finally, the clinician makes a decision for the patient according to the known information, professional knowledge and clinical experience. The decision includes three options: carry out further examination, referral and make a diagnosis.
 \textbf{b,} Currently, the AI system is hardly connected into the diagnosis process in natural clinical settings, as many important features of the clinical task are omitted. First, the AI system cannot directly handle various patients from the clinical setting, since many diseases are omitted by the AI researcher. The clinicians have to make a primary diagnosis and select the proper patients for the AI system. Second, the AI system cannot directly handle the selected patients, since lots of  patient information is omitted by the AI researcher. The clinicians have to divide the clinical diagnosis into several sub-tasks, and select a proper AI system for each specific sub-task. Third, the AI system cannot provide personalized diagnosis plan for each individual, since the AI system is usually rigid in that it can only handle the same type of data. Finally, the AI system cannot directly diagnose a patient, since the clinical value of the AI system is questionable and is in an urgent need for validation in a scientific way.
 \label{Fig2}}
\end{figure*}

\vskip 12pt
\noindent\textbf{The clinical tasks in a natural setting.} As mentioned, a natural clinical setting is open and the diagnosis itself is flexible~\cite{zhang2019landscape}, and a clinical diagnosis task is a very complex and uncertain owing to the following reasons.
\begin{itemize}
\item[a)] Affected by the subtle difference in terms of experience and knowledge of the patients, their decisions to select clinicians hold great uncertainty. As a result, the patients that have the same disease may choose different clinicians and the patients that have different diseases may go to the same clinician.
\item[b)] Comorbidity is commonly observed in a patient. For example, Caspi et al.~\cite{caspi2018all} pointed out that among individuals meeting the criteria for one psychiatric disorder in their life time, 66\% met the criteria for a second one; of those meeting criteria for two disorders, 53\% met the criteria for a third one; of those meeting the criteria for a third disorder, 41\% met criteria for a fourth one~\cite{caspi2018all}. Thus, comorbidity is an important factor that needs to be taken into account.
\item[c)] It is a consensus that many diseases have various subtypes, and different subtypes or stages of the disease may be subject to different therapeutic plans~\cite{mirnezami2012preparing}.
Thus, the final diagnosis made by the clinicians should be able to directly, accurately and duly help to select appropriate and effective therapeutic plans for the patient.
\item[d)] The workflow of a diagnosis task is flexible according to each individual situation. For example, different patients may be required to perform different examinations.
\item[e)] The data sources and types of a patient are usually multifarious and integrated, and the knowledge from different sources must be comprehensively considered~\cite{zhang2019landscape}.
\item[f)] The diagnosis is patient-centered in clinical settings. Thus, the evaluation of diagnosis must reflect the benefit of patients, such as downstream outcomes (overall survival)~\cite{parikh2019regulation}.
\end{itemize}

In a word, a clinical diagnosis task is complex and uncertain. The root cause for the current plight of the practical implementation of AI is that the essential features of clinical tasks in natural clinical settings are often omitted by AI researchers, as we discuss above.
%comorbidities need to be simultaneously considered.
\vskip 12pt
\noindent\textbf{What is wrong with the state-of-the-art AI systems in natural clinical settings.}  Currently, the clinician has to deal with massive complex and uncertain clinical tasks with their skill and experience (Fig~\ref{Fig2}a). Thus, it is highly urgent for clinicians to cooperate with AI systems to provide high-quality, high-efficiency, and low-cost healthcare solutions. Disappointedly, current AI systems show little help with regard to improving the quality of  medical care and reducing the workload of clinician due to the following reasons (Fig~\ref{Fig2}b) :
\begin{itemize}
\item[1)] Currently, the AI system is far from intelligent to be embedded in the current health system, as many important features of the clinical tasks have not been considered. The AI system only has ability to handle a few specific sub-tasks (such as the identification of medical imaging) of a clinical diagnosis carefully preprocessed by clinicians, which instead increases the workload of the clinicians (Fig~\ref{Fig2}b).
\item[2)] As a clinical diagnosis task usually consists of complex sub-tasks, the clinician needs to comprehensively consider the complete diagnosis task rather than only consider one or two specific sub-tasks, which the AI system is competent for handling. Thus, the outcome of the AI system is not appealing to the clinicians.

\item[3)] For the lack of scientific evaluation of clinical value, the AI systems are forbidden to deliver diagnosis for patients.
\end{itemize}

It is essential and significant for the AI researchers to comprehensively understand the tasks in the natural clinical settings so that they are able to improve the design and implementation of the AI systems to cope with complex and uncertain tasks in natural clinical settings. In addition, the scientific metrics and baselines are essential to evaluate the clinical value of an AI system before it is implemented in clinical settings.

\vskip 12pt
\noindent\textbf{A new direction to promote the practical implementation of AI in medicine}

\noindent Though the AI researchers claim that their AI systems have outstanding performance, their systems are still hardly implemented in natural clinical settings for ignoring the essential features of the clinical tasks. Fortunately, a clinical benchmark suite has great potential to capture the features of the clinical tasks and scientifically evaluate the performance of an AI system, thus promoting AI systems embed in healthcare system.
%thus holding great potential in addressing the aforementioned issue.
\begin{figure*}
\centering
\includegraphics[height=10cm, width=10cm]{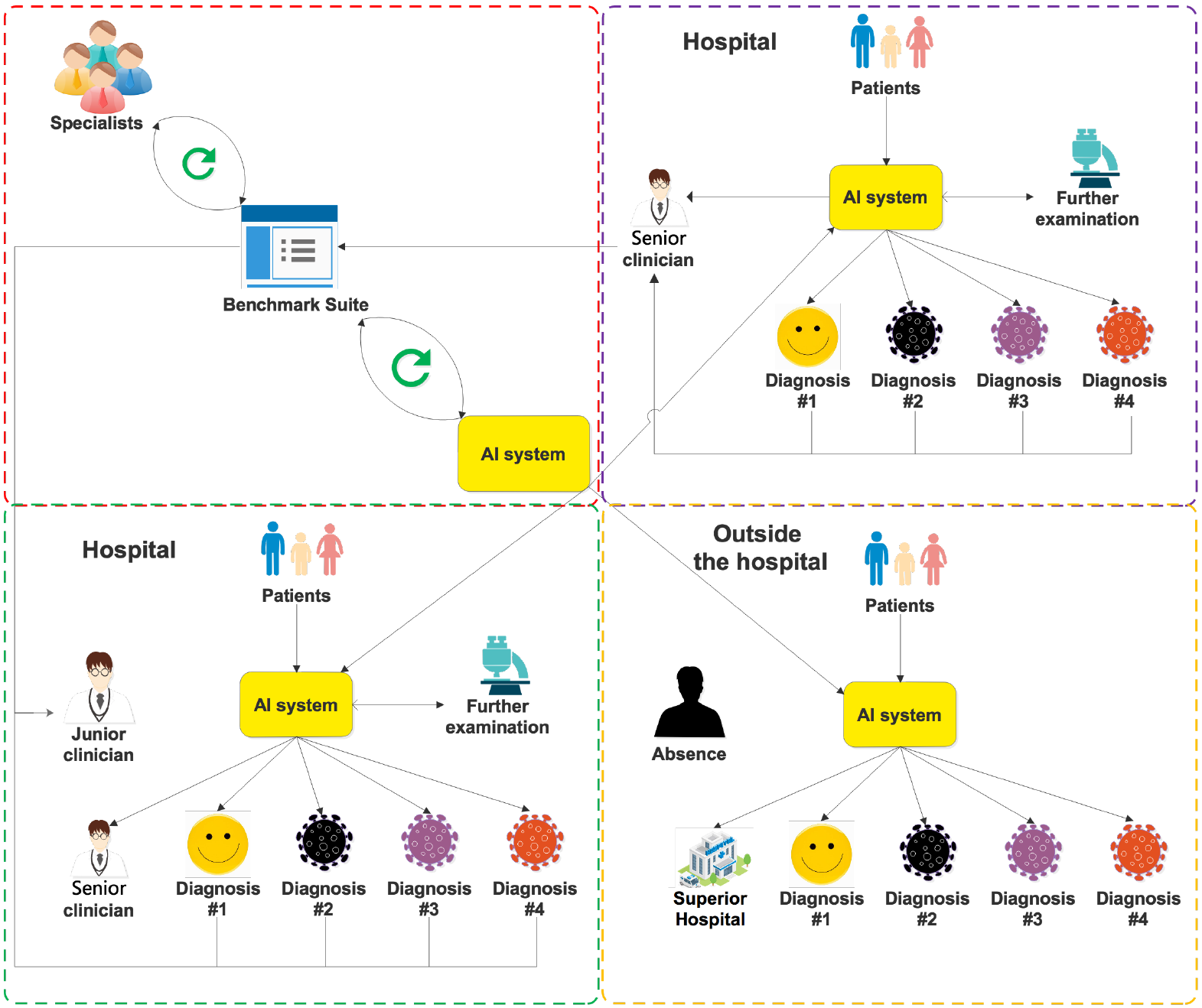}
\quad
\caption{\textbf{The future AI system in medicine.} In the future, the first thing is the cooperation between AI researchers and specialists with different backgrounds to iteratively construct a clinical benchmark suite. Secondly, the AI researchers iteratively develop and verify AI systems based on the benchmark suite. Finally, after the AI systems is approved by the regulator, they will be deployed in a natural clinical setting to provide high-quality medical care. The AI systems will play an important role in at least three classic scenarios. 1) In cooperation with senior clinicians: the AI systems will act like a clinician as shown in Fig~\ref{Fig2}a, and send the diagnostic task beyond its ability to the senior clinician. On one hand, the senior clinician reviews the diagnosis from an AI system; On the other hand, the senior clinician is in charge of the diagnostic tasks that cannot be handled by an AI system. In addition, the senior clinician can cooperate with the specialists from different backgrounds to update the benchmark suite with respect to the development of medicine and the change of the clinical settings. 2) In cooperation with junior clinicians: different from the first scenario, the junior clinicians in this scenario will cooperate with the benchmark suite to review the diagnosis of AI system, as the benchmark suite contains the `ground truth' of different patients. In this way, the benchmark suite helps the junior clinicians improve their skills.  3) Daily health management, disease prevention, chronic disease management,
out-of-hospital rehabilitation, and healthcare services in rural areas: in this scenario, the AI system will give precaution and inform the patients of seeking further medical service in hospital \label{Fig3}}
\end{figure*}

\vskip 12pt
\noindent\textbf{Benchmark suite.} In computer science, the benchmark suite is a powerful tool to simulate the workloads in the real world and evaluate the performance of the systems using the simulated workloads~\cite{fleming1986not}. On one hand, it reproduces the real workloads to help the developers comprehensively understand the real workload and improve their system design and implementation to cope with complex and uncertain workloads in the real world. On the other hand, it provides a scientific evaluation methodology and tool to objectively evaluate the system under test and guide the developer to improve the system. The benchmark suite has been successfully applied in many areas and attracted intensive research attention. Guthaus et al. ~\cite{guthaus2001mibench} presented a benchmark suite for the CPU evaluation. Wang et al.~\cite{wang2014bigdatabench} presented a comprehensive big data benchmark suite, named BigDataBench, for the architecture and system evaluation. Geiger et al.~\cite{geiger2012we} constructed a vision benchmark suite, named KITTI, for autonomous driving evaluation. Thus, we strongly believe that the clinical AI benchmark suite holds the potential to promote the implementation of AI systems in natural clinical settings.

\vskip 12pt
\noindent\textbf{The future of an AI system with a clinical benchmark suite.} Inspired by the successes of AI in
various realms, we are optimistic that AI has potential to improve the quality of healthcare. However, as the specialty of healthcare  (involve our health), we believe that AI system should be harmoniously embedded in the current healthcare systems to cooperate with clinicians to provide high quality and low cost of medical service instead of replacing clinicians. And, it is the key to help the AI researchers capture the features of the clinical tasks, improve the design of the AI system and optimize the implementation of the AI system, and promote the AI systems to be harmoniously embedded in the current healthcare system.

In the future, the clinical benchmark needs to be paid more attention, as it is expected to play a significant role in the development of AI system. The clinical benchmark will run throughout the entire lifecycle of the AI system in medicine (Fig~\ref{Fig3}) undertaking the following tasks: (1) help the researchers understand the natural clinical settings and clinical tasks before the implementation of an AI system, and finally promote the AI system to implement in a natural clinical setting, (2) comprehensively test the performance of the AI system in a simulated clinical setting in the development process, (3) provide evaluation indicators and baselines for the regulatory authorities. In this way, AI system can be embedded in the current healthcare system. In addition, the AI system will be able to \textbf{A)} reduce the workload of clinicians by automatically handling essential clinician sub-tasks, \textbf{B)} improve the skill of junior clinician by providing a simulated clinical setting, \textbf{C)} give precaution and inform the patients of further medicine service in hospital with regard to daily health management, disease prevention,  chronic disease management, and out-of-hospital rehabilitation, and \textbf{D)} provide preliminary healthcare service in the rural areas.
%So that the AI system has the ability to embed into the current healthcare system. And, the

\begin{figure*}
\centering
\includegraphics[height=6cm, width=10cm]{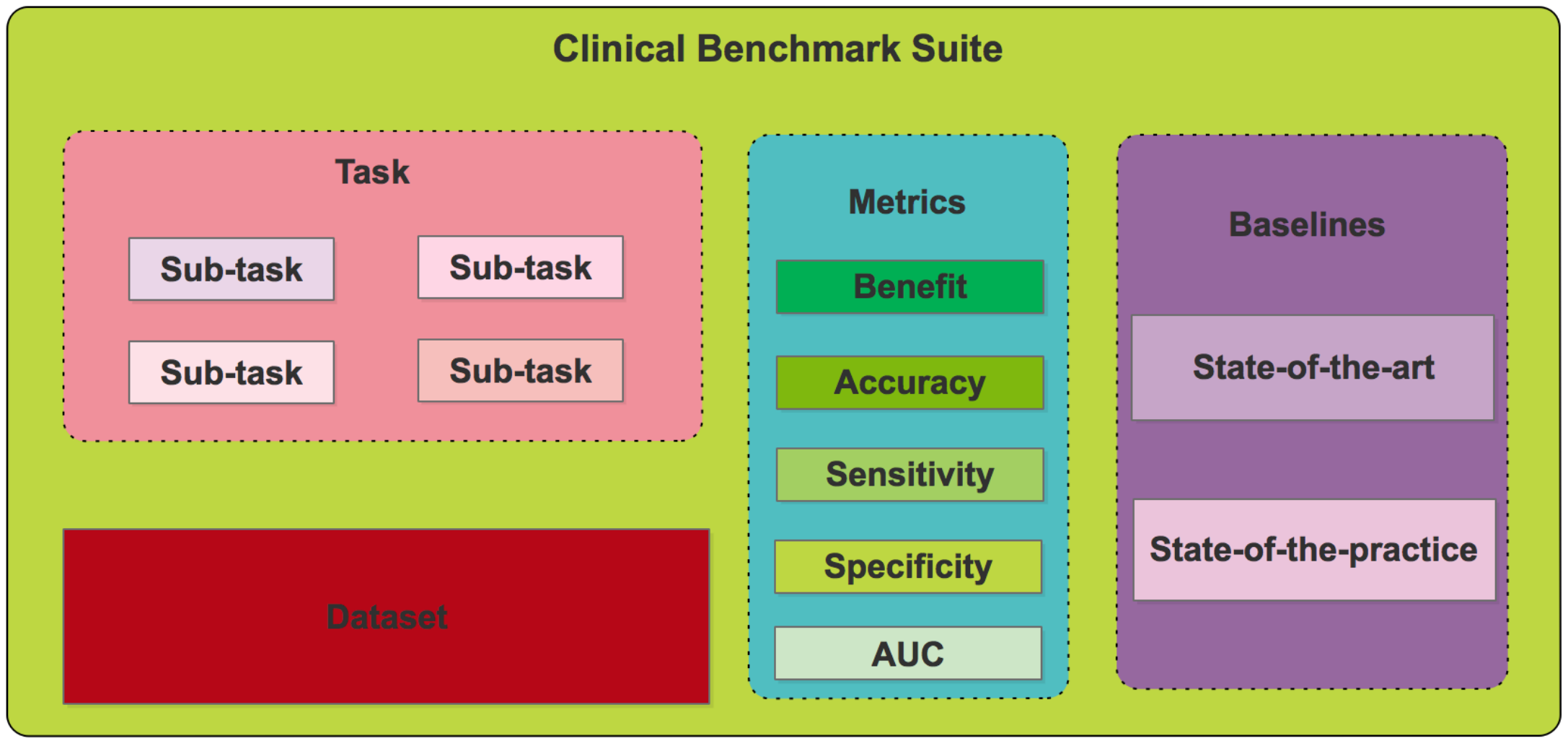}
\quad
\caption{\textbf{The clinical benchmark suite.} The benchmark suite consists of data sets, tasks, metrics and baselines. The data sets contain the patient information from the real clinical practice. A clinical diagnosis task is  a flexible diagnosis process consisting of several sub-tasks. The metrics consists of the patient benefit metrics and the other common performance metrics. The baselines contain both state-of-the-art and state-of-the-practice methods. \label{Fig4}}
\end{figure*}

\vskip 12pt
\noindent\textbf{The key issues of the research on clinical benchmarks.} In the realm of medicine, the clinical benchmark suite is a new conception. Thus, there are many key issues of the clinical benchmark suite that must be investigated (Fig~\ref{Fig4}).
\begin{itemize}
\item[a)]\textbf{Data sets.} Unlike the carefully-constructed data set for the current AI system, the data set in the real clinical practice should contain all the diverse information (including so-called bad data), which is conducive to the more accurate restoring of the natural clinical settings and tasks.
\item[b)]\textbf{Tasks.} The task is the workload of the AI system in a natural clinical setting, where each clinical diagnosis task is personalized and flexible according to each individual situation. And the clinical diagnosis is continuous until the result of the diagnosis is able to effectively guide the treatment.  For most diseases there is no one-size-fits-all diagnosis task, and each diagnosis task usually consists of several different sub-tasks according to each individual patient. 
\item[c)]\textbf{Metrics.} The purpose of the diagnosis is to obtain the evidence to assist the clinicians in making the suitable and effective medical treatment plan for the patients.
Thus, the metrics must reflect the benefit of the selected treatment for the patient. However, the current metrics of AI system (such as the accuracy of prediction) do not have such capabilities. Another important issue is that current metrics cannot represent the standards although they can be used as a reference~\cite{park2018methodologic, parikh2019regulation}.
\item[d)]\textbf{Baselines.} The benchmark suite must include two baselines. One is the state-of-the-art baseline that is used to evaluate the AI system from a research technical perspective. The other is the state-of-the-practice baseline that is used by the clinicians in the natural clinical settings. This baseline is used to evaluate the clinical value of the AI system from a perspective of clinical practice.
\end{itemize}

\vskip 12pt
\noindent\textbf{Conclusion}

AI systems are currently witnessing excessive hype of their `super power' in medicine research and thus capturing intensive attention of the public, which leads to the increasing expectation of public to used AI technologies to improve healthcare. However, current AI technologies are far from being mature to be implemented in natural clinical settings, in spite of their rapid advancement~\cite{Jianxing2019}. And, many clinicians  gradually losing faith in AI~\cite{ brocklehurst2017computerised,chen2017machine,mittelman2018patient,maddox2019questions}.  It is worth noting that the increasing gap between the high expectation and unsatisfactory practical implementation of AI technology is one of the main reasons why AI fell into the first trough.
%In addition, it is worth noting that the increasing gap between the practical implementation and expectation of the AI is one of the main reasons why AI fell into the first trough. 
Hence, it is imperative to actively promote the practical implementation of AI technology and objectively evaluate the current AI technology. Inspired by the success of the benchmark in the computer communities, we strongly believe that the clinical benchmark is a powerful tool to promote the AI technology to be effectively applied in the real clinical practice, and provide an objective evaluation of the current AI technologies. The clinical benchmark suite holds a promising future as a new direction to promote the implementation of AI in clinical practice. Alike the animal models that play an important role in the drug development process, a clinical benchmark suite will also play an important role spanning the whole lifecycle of an AI system for clinical tasks not only in the design, implementation, and evaluation, but also in a natural clinical setting.

%The clinical benchmark will be a promising direction to promote the practical implementation of the artificial intelligence in clinical practice. Just as the animal models play an important role in drug development, a clinical benchmark suite will  play an important role not only during the design, implementation, and evaluation of an AI system, but also in a natural clinical setting.

%For the excessive hype of the `super power' of the AI system in medicine researches, the AI research on the medicine has captured massive attention of the public, and the masses are increasingly expecting AI to improve healthcare. However, although the AI technologies are rapidly advancing, their implementation into natural clinical settings has not yet mature~\cite{Jianxing2019}. The clinicians are gradually losing faith in AI~\cite{ brocklehurst2017computerised,chen2017machine,mittelman2018patient,maddox2019questions}. 

\vskip 12pt
\vskip 12pt
\vskip 12pt
\vskip 12pt

%% Put the bibliography here, most people will use BiBTeX in
%% which case the environment below should be replaced with
%% the \bibliography{} command.
%\bibliography{reference}

%\begin{thebibliography}{10}

%\end{thebibliography}

%% Here is the endmatter stuff: Supplementary Info, etc.
%% Use \item's to separate, default label is "Acknowledgements"

\begin{addendum}
 \item This work is supported by the Major Program of National Natural Science Foundation of China (Grant No. 61432006).
  \item[Author contributions] Y.Y.H., Z.F.Z., N.N.W. and N.Q.L. conceived and designed the article, performed the literature review and wrote and revised the manuscript.  J.F.Z. and Z.F.Z. supervised the work. M.J.D. organized the data and revised the manuscript. T.S.H. revised the manuscript.
 \item[Competing Interests] The authors declare no competing interests.
 \item[Correspondence] Correspondence and requests for materials
should be addressed to Institute of Computing Technology, Chinese Academy of Sciences, No. 6 South Road of Academy of Sciences, Haidian District, Beijing 100089, China.~(email: zhifeiz@ccmu.edu.cn or zhanjianfeng@ict.ac.cn).
\end{addendum}

%%
%% TABLES
%%
%% If there are any tables, put them here.
%%
%\end{multicols}
\end{document}